# Speech Recognition by Machine: A Review

M.A.Anusuya

Department of Computer Science and Engineering
Sri Jaya chamarajendra College of Engineering
Mysore, India
.

S.K.Katti

Department of Computer Science and Engineering
Sri Jayachamarajendra College of Engineering
Mysore, India
.

*Abstract*— This paper presents a brief survey on Automatic Speech Recognition and discusses the major themes and advances made in the past 60 years of research, so as to provide a technological perspective and an appreciation of the fundamental progress that has been accomplished in this important area of speech communication. After years of research and development the accuracy of automatic speech recognition remains one of the important research challenges (eg., variations of the context, speakers, and environment).The design of Speech Recognition system requires careful attentions to the following issues: Definition of various types of speech classes, speech representation, feature extraction techniques, speech classifiers, database and performance evaluation. The problems that are existing in ASR and the various techniques to solve these problems constructed by various research workers have been presented in a chronological order. Hence authors hope that this work shall be a contribution in the area of speech recognition. The objective of this review paper is to summarize and compare some of the well known methods used in various stages of speech recognition system and identify research topic and applications which are at the forefront of this exciting and challenging field.

## I. INTRODUCTION

### A. Definition of speech recognition:

Speech Recognition (is also known as Automatic Speech Recognition (ASR), or computer speech recognition) is the process of converting a speech signal to a sequence of words, by means of an algorithm implemented as a computer program.

### 1.2 Basic Model of Speech Recognition:

Research in speech processing and communication for the most part, was motivated by people's desire to build mechanical models to emulate human verbal communication capabilities. Speech is the most natural form of human communication and speech processing has been one of the most exciting areas of the signal processing. Speech recognition technology has made it possible for computer to follow human voice commands and understand human languages. The main goal of speech recognition area is to develop techniques and systems for speech input to machine. Speech is the primary means of communication between humans. For reasons ranging from technological curiosity about the mechanisms for mechanical realization of human speech capabilities to desire to automate simple tasks which necessitates human machine interactions and research in automatic speech recognition by machines has attracted a great deal of attention for sixty years[76]. Based on major advances in statistical modeling of speech, automatic speech recognition systems today find widespread application in tasks that require human machine interface, such as automatic call processing in telephone networks, and query based information systems that provide updated travel information, stock price quotations, weather reports, Data entry, voice dictation, access to information: travel, banking, Commands, Avoinics, Automobile portal, speech transcription, Handicapped people (blind people) supermarket, railway reservations etc. Speech recognition technology was increasingly used within telephone networks to automate as well as to enhance the operator services. This report reviews major highlights during the last six decades in the research and development of automatic speech recognition, so as to provide a technological perspective. Although many technological progresses have been made, still there remains many research issues that need to be tackled.

Fig.1 shows a mathematical representation of speech recognition system in simple equations which contain front end unit, model unit, language model unit, and search unit. The recognition process is shown below (Fig .1).

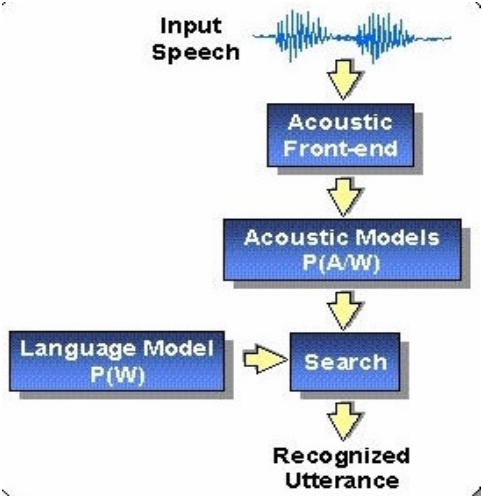

*Fig.1   Basic model of speech recognition*

The standard approach to large vocabulary continuous speech recognition is to assume a simple probabilistic model of speech production whereby a specified word sequence, W, produces an acoustic observation sequence Y, with probability P(W,Y). The goal is then to decode the word string, based on the acoustic observation sequence, so that the decoded string has the maximum a posteriori (MAP) probability.

$$\hat{P}(W/A)= \arg \max_{W} P(W/A) \qquad \dots \dots (1)$$

Using Baye's rule, equation   (1) can be written as

$$P(W/A)=\frac{P(A/W)P(W)}{P(A)} \qquad \dots \dots (2)$$

Since P(A) is independent of W, the MAP decoding rule of equation(1) is

$$\hat{W}=\arg\max_{w} \hat{P}(A/W)P(W) \qquad \dots \dots (3)$$

The first term in equation (3) P(A/W), is generally called the acoustic model, as it estimates the probability of a sequence of acoustic observations, conditioned on the word string. Hence P(A/W) is computed. For large vocabulary speech recognition systems, it is necessary to build statistical models for sub word speech units, build up word models from these sub word speech unit models (using a lexicon to describe the composition of words), and then postulate word sequences and evaluate the acoustic model probabilities via standard concatenation methods. The second term in equation (3) P(W), is  called the language model.  It describes the probability associated with a postulated sequence of words. Such language models can incorporate both syntactic and semantic constraints of the language and the recognition task.

## 1.3 Types of Speech Recognition

Speech recognition systems can be separated in several different classes by describing what types of utterances they have the ability to recognize. These classes are classified as the following:

**Isolated Words**:
Isolated word recognizers usually require each utterance to have quiet (lack of an audio signal) on both sides of the sample window. It accepts single words or single utterance at a time. These systems have "Listen/Not-Listen" states, where they require the speaker to wait between utterances (usually doing processing during the pauses). Isolated Utterance might be a better name for this class.

**Connected Words:**
Connected word systems (or more correctly 'connected utterances') are similar to isolated words, but allows separate utterances to be 'run-together' with a minimal pause between them.

**Continuous Speech:**
Continuous speech recognizers allow users to speak almost naturally, while the computer determines the content. (Basically, it's computer dictation). Recognizers with continuous speech capabilities are some of the most difficult to create because they utilize special methods to determine utterance boundaries.

**Spontaneous Speech:**
At a basic level, it can be thought of as speech that is natural sounding and not rehearsed. An ASR system with spontaneous speech ability should be able to handle a variety of natural speech features such as words being run together, "ums" and "ahs", and even slight stutters.

## 1.4 Applications of Speech Recognition:

Various applications of speech recognition domain have been discussed in the following table 1.

Table 1: Applications of speech recognition:

| Problem Domain | Application | Input pattern | Pattern classes |
|---|---|---|---|
| Speech/Telehphone/ Communication Sector/Recognition | Telephone directory enquiry without operator assistance | Speech wave form | Spoken words |
| Education Sector | Teaching students of foreign languages to pronounce vocabulary correctly. Teaching overseas students to pronounce English correctly. | Speech wave form | Spoken words |
| | Enabling students who are physically handicapped and unable to use a keyboard to enter text verbally

Narrative oriented research, where transcripts are automatically generated. This would remove the time to manually generate the transcript, and human error. | | |
| Outside education sector | Computer and video games, Gambling, Precision surgery | Speech wave form | Spoken words |
| Domestic sector | Oven, refrigerators, dishwashers and washing machines | Speech wave form | Spoken words |
| Military sector | High performance fighter aircraft, Helicopters, Battle management, Training air traffic controllers, Telephony and other domains, people with disabilities | Speech wave form | Spoken words |
| Artificial Intelligence sector | Robotics | Speech wave form | Spoken words |
| Medical sector | Health care, Medical Transcriptions (digital speech to text) | Speech wave form | Spoken words |
| General: | Automated transcription, Telematics, Air traffic control, Multimodal interacting, court reporting, Grocery shops | Speech wave form | Spoken words |
| Physically Handicapped | Useful to the people with limited mobility in their arms and hands or for those with sight | Speech wave form | Spoken words |
| Dictation | Dictation systems on the market accepts continuous speech input which replaces menu system. | Speech wave form | Spoken words |
| Translation | It is an advanced application which translates from one language to another. | Speech wave form | Spoken words |

## 1.5 Automatic Speech Recognition system classification:

The following tree structure emphasizes the speech processing applications. Depending on the chosen criterion, Automatic Speech Recognition systems can be classified as shown in figure 2.

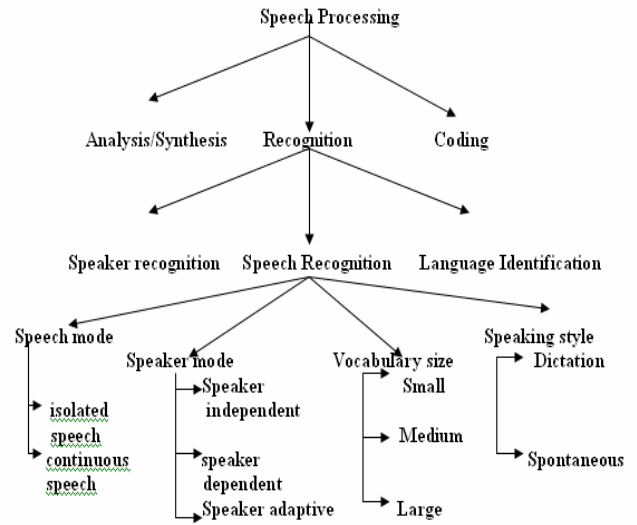

Fig 2 Speech processing classification

## 1.6 Relevant issues of ASR design:

Main issues on which recognition accuracy depends have been presented in the table 2.

Table 2: Relevant issues of ASR design

| Environment | Type of noise; signal/noise ratio; working conditions |
|---|---|
| Transducer | Microphone; telephone |
| Channel | Band amplitude; distortion; echo |
| Speakers | Speaker dependence/independence Sex, Age; physical and psychical state |
| Speech styles | Voice tone(quiet, normal, shouted); Production(isolated words or continuous speech read or spontaneous speech) Speed(slow, normal, fast) |
| Vocabulary | Characteristics of available training data; specific or generic vocabulary; |

## 2. Approaches to speech recognition:

Basically there exist three approaches to speech recognition. They are

- Acoustic Phonetic Approach
- Pattern Recognition Approach
- Artificial Intelligence Approach

### 2.1 Acoustic phonetic approach:

The earliest approaches to speech recognition were based on finding speech sounds and providing appropriate labels to these sounds. This is the basis of the acoustic phonetic approach (Hemdal and Hughes 1967), which postulates that there exist finite, distinctive phonetic units (phonemes) in spoken language and that these units are broadly characterized by a set of acoustics properties that are manifested in the speech signal over time. Even though, the acoustic properties of phonetic units are highly variable, both with speakers and with neighboring sounds (the so-called co articulation effect), it is assumed in the acoustic-phonetic approach that the rules governing the variability are straightforward and can be readily learned by a machine. The first step in the acoustic phonetic approach is a spectral analysis of the speech combined with a feature detection that converts the spectral measurements to a set of features that describe the broad acoustic properties of the different phonetic units. The next step is a segmentation and labeling phase in which the speech signal is segmented into stable acoustic regions, followed by attaching one or more phonetic labels to each segmented region, resulting in a phoneme lattice characterization of the speech. The last step in this approach attempts to determine a valid word (or string of words) from the phonetic label sequences produced by the segmentation to labeling. In the validation process, linguistic constraints on the task (i.e., the vocabulary, the syntax, and other semantic rules) are invoked in order to access the lexicon for word decoding based on the phoneme lattice. The acoustic phonetic approach has not been widely used in most commercial applications ([76], Refer fig.2.32. p.81).The following table 3 broadly gives the different speech recognition techniques.

Table 3: Speech Recognition Techniques

| Approach | Representation | Recognition Function | Typical Criterion |
|---|---|---|---|
| Acoustic phonetic approach | Phonemes/segmentation And labeling | Probabilistic lexical access procedure | Log likelihood ratio |
| Pattern recognition approach | | | |
| • Template | Speech samples, pixels & curves | Correlation, distance measure | Classification error |
| • DTW | Set of a sequences of spectral vectors | Dynamic warping | Dissimilarity measure |
| • VQ | Set of spectral vectors | optimal algorithm | Euclidian distance |
| • Statistical | Features | Clustering functions (code book) Discriminant functions | Classification error |
| Neural network | Speech features/perceptrons/ Rules\units\procedures | Network function | Mean square error |
| Support vector machine | Kernel based features | Maximal margin hyperplane, Radial basis function classifier(fitting functions) | Minimizing a bound on the Generalization error. |
| Artificial Intelligence approach | Knowledge based | | Word error probability |

### 2.2 Pattern Recognition approach:

The pattern-matching approach (Itakura 1975; Rabiner 1989; Rabiner and Juang 1993) involves two essential steps— namely, pattern training and pattern comparison. The essential feature of this approach is that it uses a well formulated mathematical framework and establishes consistent speech pattern representations, for reliable pattern comparison, from a set of labeled training samples via a formal training algorithm. A speech pattern representation can be in the form of a speech template or a statistical model (e.g., a HIDDEN MARKOV MODEL or HMM) and can be applied to a sound (smaller than a word), a word, or a phrase. In the pattern-comparison stage of the approach, a direct comparison is made between the unknown speeches (the speech to be recognized) with each possible pattern learned in the training stage in order to determine the identity of the unknown according to the goodness of match of the patterns. The pattern-matching approach has become the predominant method for speech recognition in the last six decades ([76] Refer fig.2.37. pg.87). A block schematic diagram of pattern recognition is presented in fig.3 below. In this, there exists two methods namely template approach and stochastic approach.

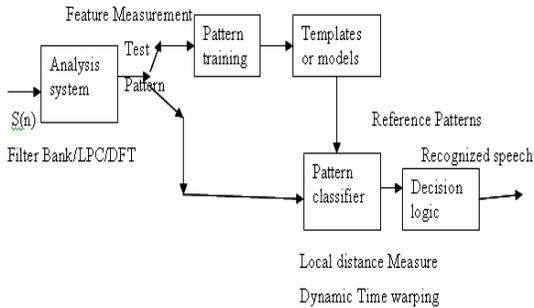

Fig 3. Block diagram of Pattern recognition speech recognizer

## 2.2.1. Template Based Approach:

Template based approach [97] to speech recognition have provided a family of techniques that have advanced the field considerably during the last six decades. The underlying idea is simple. A collection of prototypical speech patterns are stored as reference patterns representing the dictionary of candidate's words. Recognition is then carried out by matching an unknown spoken utterance with each of these reference templates and selecting the category of the best matching pattern. Usually templates for entire words are constructed. This has the advantage that, errors due to segmentation or classification of smaller acoustically more variable units such as phonemes can be avoided. In turn, each word must have its own full reference template; template preparation and matching become prohibitively expensive or impractical as vocabulary size increases beyond a few hundred words. One key idea in template method is to derive a typical sequences of speech frames for a pattern(a word) via some averaging procedure, and to rely on the use of local spectral distance measures to compare patterns. Another key idea is to use some form of dynamic programming to temporarily align patterns to account for differences in speaking rates across talkers as well as across repetitions of the word by the same talker.

## 2.2.2. Stochastic Approach:

Stochastic modeling [97] entails the use of probabilistic models to deal with uncertain or incomplete information. In speech recognition, uncertainty and incompleteness arise from many sources; for example, confusable sounds, speaker variability's, contextual effects, and homophones words. Thus, stochastic models are particularly suitable approach to speech recognition. The most popular stochastic approach today is hidden Markov modeling. A hidden Markov model is characterized by a finite state markov model and a set of output distributions. The transition parameters in the Markov chain models, temporal variabilities, while the parameters in the output distribution model, spectral variabilities. These two types of variabilities are the essence of speech recognition.

Compared to template based approach, hidden Markov modeling is more general and has a firmer mathematical foundation. A template based model is simply a continuous density HMM, with identity covariance matrices and a slope constrained topology. Although templates can be trained on fewer instances, they lack the probabilistic formulation of full HMMs and typically underperforms HMMs. Compared to knowledge based approaches; HMMs enable easy integration of knowledge sources into a compiled architecture. A negative side effect of this is that HMMs do not provide much insight on the recognition process. As a result, it is often difficult to analyze the errors of an HMM system in an attempt to improve its performance. Nevertheless, prudent incorporation of knowledge has significantly improved HMM based systems.

## 2.3. Dynamic Time Warping(DTW):

**Dynamic time warping** is an algorithm for measuring similarity between two sequences which may vary in time or speed. For instance, similarities in walking patterns would be detected, even if in one video, the person was walking slowly and if in another, he or she were walking more quickly, or even if there were accelerations and decelerations during the course of one observation. DTW has been applied to video, audio, and graphics — indeed, any data which can be turned into a linear representation can be analyzed with DTW. A well known application has been automatic speech recognition, to cope with different speaking speeds. In general, DTW is a method that allows a computer to find an optimal match between two given sequences (e.g. time series) with certain restrictions. The sequences are "warped" non-linearly in the time dimension to determine a measure of their similarity independent of certain non-linear variations in the time dimension. This sequence alignment method is often used in the context of hidden Markov models.

One example of the restrictions imposed on the matching of the sequences is on the monotonicity of the mapping in the time dimension. Continuity is less important in DTW than in other pattern matching algorithms; DTW is an algorithm particularly suited to matching sequences with missing information, provided there are long enough segments for matching to occur. The optimization process is performed using dynamic programming, hence the name.

## 2.4. Vector Quantization(VQ):

Vector Quantization(VQ)[97] is often applied to ASR. It is useful for speech coders, i.e., efficient data reduction. Since transmission rate is not a major issue for ASR, the utility of VQ here lies in the efficiency of using compact codebooks for reference models and codebook searcher in place of more costly evaluation methods. For IWR, each vocabulary word gets its own VQ codebook, based on training sequence of several repetitions of the word. The test speech is evaluated by all codebooks and ASR chooses the word whose codebook yields the lowest distance measure. In basic VQ, codebooks have no explicit time information (e.g., the temporal order of phonetic segments in each word and their relative durations are ignored), since codebook entries are not ordered and can come from any part of the training words. However, some indirect durational cues are preserved because the codebook entries are chosen to minimize average distance across all training frames, and frames, corresponding to longer acoustic segments ( e.g., vowels) are more frequent in the training data. Such segments are thus more likely to specify code words than less frequent consonant frames, especially with small codebooks. Code words nonetheless exist for constant frames because such frames would otherwise contribute large frame distances to the codebook. Often a few code words suffice to represent many frames during relatively steady sections of vowels, thus allowing more codeword to represent short, dynamic portions of the words. This relative emphasis that VQ puts on speech transients can be an advantage over other ASR comparison methods for vocabularies of similar words.

### 2.5. Artificial Intelligence approach (Knowledge Based approach)

The Artificial Intelligence approach [97] is a hybrid of the acoustic phonetic approach and pattern recognition approach. In this, it exploits the ideas and concepts of Acoustic phonetic and pattern recognition methods. Knowledge based approach uses the information regarding linguistic, phonetic and spectrogram. Some speech researchers developed recognition system that used acoustic phonetic knowledge to develop classification rules for speech sounds. While template based approaches have been very effective in the design of a variety of speech recognition systems; they provided little insight about human speech processing, thereby making error analysis and knowledge-based system enhancement difficult. On the other hand, a large body of linguistic and phonetic literature provided insights and understanding to human speech processing. In its pure form, knowledge engineering design involves the direct and explicit incorporation of expert's speech knowledge into a recognition system. This knowledge is usually derived from careful study of spectrograms and is incorporated using rules or procedures. Pure knowledge engineering was also motivated by the interest and research in expert systems. However, this approach had only limited success, largely due to the difficulty in quantifying expert knowledge. Another difficult problem is the integration of many levels of human knowledge − phonetics, phonotactics, lexical access, syntax, semantics and pragmatics. Alternatively, combining independent and asynchronous knowledge sources optimally remains an unsolved problem. In more indirect forms, knowledge has also been used to guide the design of the models and algorithms of other techniques such as template matching and stochastic modeling. This form of knowledge application makes an important distinction between knowledge and algorithms − Algorithms enable us to solve problems. Knowledge enable the algorithms to work better. This form of knowledge based system enhancement has contributed considerably to the design of all successful strategies reported. It plays an important role in the selection of a suitable input representation, the definition of units of speech, or the design of the recognition algorithm itself.

### 2.6. Connectionist Approaches (Artificial Neural Networks):

The artificial intelligence approach ( [97], Lesser et al. 1975; Lippmann 1987) attempts to mechanize the recognition procedure according to the way a person applies intelligence in visualizing, analyzing, and characterizing speech based on a set of measured acoustic features. Among the techniques used within this class of methods are use of an expert system (e.g., a neural network) that integrates phonemic, lexical, syntactic, semantic, and even pragmatic knowledge for segmentation and labeling, and uses tools such as artificial NEURAL NETWORKS for learning the relationships among phonetic events. The focus in this approach has been mostly in the representation of knowledge and integration of knowledge sources. This method has not been widely used in commercial systems. Connectionist modeling of speech is the youngest development in speech recognition and still the subject of much controversy. In connectionist models, knowledge or constraints are not encoded in individual units, rules, or procedures, but distributed across many simple computing units. Uncertainty is modeled not as likelihoods or probability density functions of a single unit, but by the pattern of activity in many units. The computing units are simple in nature, and knowledge is not programmed into any individual unit function; rather, it lies in the connections and interactions between linked processing elements. Because the style of computation that can be performed by networks of such units bears some resemblance to the style of computation in the nervous system. Connectionist models are also referred to as neural networks or artificial neural networks. Similarly, parallel distributed processing or massively distributed processing are terms used to describe these models. Not unlike stochastic models, connectionist models rely critically on the availability of good training or learning strategies. Connectionist learning seeks to optimize or organize a network of processing elements. However, connectionist models need not make assumptions about the underlying probability distributions. Multilayer neural networks can be trained to generate rather complex nonlinear classifiers or mapping function. The simplicity and uniformity of the underlying processing element makes connectionist models attractive for hardware implementation, which enables the operation of a net to be simulated efficiently. On the other hand, training often requires much iteration over large amounts of training data, and can, in some cases, be prohibitively expensive. While connectionism appears to hold great promise as plausible model of cognition, may question relating to the concrete realization of practical connectionist recognition techniques, still remain to be resolved.

### 2.7. Support Vector Machine(SVM):

One of the powerful tools for pattern recognition that uses a discriminative approach is a SVM[97]. SVMs use linear and nonlinear separating hyper-planes for data classification. However, since SVMs can only classify fixed length data vectors, this method cannot be readily applied to task involving variable length data classification. The variable length data has to be transformed to fixed length vectors before SVMs can be used. It is a generalized linear classifier with maximum-margin fitting functions. This fitting function provides regularization which helps the classifier generalized better. The classifier tends to ignore many of the features. Conventional statistical and Neural Network methods control model complexity by using a small number of features ( the problem dimensionality or the number of hidden units). SVM controls the model complexity by controlling the VC dimensions of its model. This method is independent of dimensionality and can utilize spaces of very large dimensions spaces, which  permits a construction of very large number of non-linear features and then performing "adaptive feature selection"  during training. By shifting all non-linearity to the features, SVM can use linear model for which VC dimensions is known. For example, a support vector machine can be used as a regularized radial basis function classifier.

**2.8. Taxonomy of Speech Recognition:**

Existing techniques for speech recognition have been represented diagrammatically in the following figure 4.

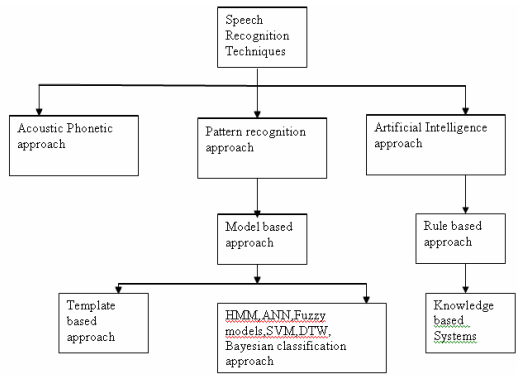

Fig 4. Taxonomy of speech recognition

**3. Feature Extraction:**
In speech recognition, the main goal of the feature extraction step is to compute a parsimonious sequence of feature vectors providing a compact representation of the given input signal. The feature extraction is usually performed in three stages. The first stage is called the speech analysis or the acoustic front end. It performs some kind of spectro temporal analysis of the signal and generates raw features describing the envelope of the power spectrum of short speech intervals. The second stage compiles an extended feature vector composed of static and dynamic features. Finally, the last stage( which is not always present) transforms these extended feature vectors into more compact and robust vectors that are then supplied to the recognizer. Although there is no real consensus as to what the optimal feature sets should look like, one usually would like them to have the following properties: they should allow an automatic system to discriminate between different through similar sounding speech sounds, they should allow for the automatic creation of acoustic models for these sounds without the need for an excessive amount of training data, and they should exhibit statistics which are largely invariant across speakers and speaking environment.

**3.1.Various methods for Feature Extraction in speech recognition are broadly shown in the following table 4.**

Table 4: feature extraction methods

| Method | Property | Comments |
|---|---|---|
| Principal Component Analysis(PCA) | Non linear feature extraction method, Linear map; fast; eigenvector-based | Traditional, eigenvector based method, also known as karhuneu-Loeve expansion; good for Gaussian data. |
| Linear Discriminant Analysis(LDA) | Non linear feature extraction method, Supervised linear map; fast; eigenvector-based | Better than PCA for classification; |
| Independent Component Analysis (ICA) | Non linear feature extraction method, Linear map, iterative non-Gaussian | Blind course separation, used for de-mixing non-Gaussian distributed sources(features) |
| Linear Predictive coding | Static feature extraction method,10 to 16 lower order co-efficient, | |
| Cepstral Analysis | Static feature extraction method, Power spectrum | Used to represent spectral envelope |
| Mel-frequency scale analysis | Static feature extraction method, Spectral analysis | Spectral analysis is done with a fixed resolution along a subjective frequency scale i.e. Mel-frequency scale. |
| Filter bank analysis | Filters tuned required frequencies | |
| Mel-frequency cepstrum (MFFCs) | Power spectrum is computed by performing Fourier Analysis | |
| Kernel based feature extraction method | Non linear transformations, | Dimensionality reduction leads to better classification and it is used to |
| | | remove noisy and redundant features, and improvement in classification error |
| Wavelet | Better time resolution than Fourier Transform | It replaces the fixed bandwidth of Fourier transform with one proportional to frequency which allow better time resolution at high frequencies than Fourier Transform |
| Dynamic feature extractions i)LPC ii)MFCCs | Acceleration and delta coefficients i.e. II and III order derivatives of normal LPC and MFCCs coefficients | |
| Spectral subtraction | Robust Feature extraction method | |
| Cepstral mean subtraction | Robust Feature extraction | |
| RASTA filtering | For Noisy speech | |
| Integrated Phoneme subspace method | A transformation based on PCA+LDA+ICA | Higher Accuracy than the existing methods |

## 4. Classifiers [149]:

In speech recognition a supervised pattern classification system is trained with labeled examples; that is, each input pattern has a class label associated with it. Pattern classifiers can also be trained in an unsupervised fashion. For example in a technique known as vector quantization, some representation of the input data is clustered by finding implicit groupings in the data. The resulting table of cluster centers is known as a codebook, which can be used to index new vectors by finding the cluster center that is closest to the new vectors. For the case of speech, fig.4a. shows an extreme case of some vowels represented by their formant frequencies F1 and F2. The vowels represented are, as pronounced in the words bot(/a/) and boot (/u/). Notice that they fall into nice groupings.

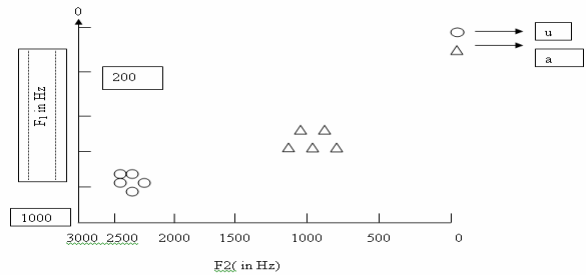

Fig.4a. Classification example

Once a feature selection or classification procedure finds a proper representation, a classifier can be designed using a number of possible approaches. In practice, the choice of a classifier is a difficult problem and it is often based on which classifier(s) happen to be available, or best known, to the user. The three different approaches are identified to design a classifier. The simplest and the most intuitive approach to classifier design is based on the concept of similarity: patterns that are similar should be assigned to the same class. So, once a good metric has been established to define similarity, patterns can be classified by template matching or the minimum distance classifier using a few prototypes per class. The choice of the metric and the prototypes is crucial to the success of this approach. In the nearest mean classifier, selecting prototypes is very simple and robust; each pattern class is represented by a single prototype which is the mean vector of all the training patterns in that class. More advanced techniques for computing prototypes are vector quantization [154, 155] and Learning Vector Quantization [156], and the data reduction methods associated with the one-nearest neighbor decision rule (1-NN) such as editing and condensing [157]. The most straightforward 1-NN rule can be conveniently used as a benchmark for all the other classifiers since it appears to always provide a reasonable classification performance in most applications. Further, as the 1-NN classifier does not require any user-specified parameters (except perhaps the distance metric used to find the nearest neighbor, but Euclidean distance is commonly used), its classification results are implementation independent. In many classification problems, the classifier is expected to have some desired invariant properties. An example is the shift invariance of characters in character recognition, a change in a character's location should not affect its classification. If the pre-processing or the representation scheme does not normalize the input pattern for this invariance, then the same character may be represented at multiple positions in the feature space. These positions define a one-dimensional subspace. As more invariants are considered, the dimensionality of this subspace correspondingly increases. Template matching or the nearest mean classifier can be viewed as finding the nearest subspace [158]. The second main concept used for designing pattern classifiers is based on the probabilistic approach. The optimal Bayes decision rule (with the 0/1 loss function) assigns a pattern to the class with the maximum posterior probability. This rule can be modified to take into account, costs associated with different types of classifications. For known class conditional densities, the Bayes decision rule gives the optimum classifier, in the sense that for given prior probabilities, loss function and class-conditional densities, no other decision rule will have a lower risk (i.e., expected value of the loss function, for example, probability of error). If the prior class probabilities are equal and a 0/1 loss function, the Bayes decision rule and the maximum likelihood decision rule exactly coincide. In practice, the empirical Bayes decision rule, or 'plug-in' rule, is used. The estimates of the densities are used in place of the true densities. These density estimates are either parametric or nonparametric. Commonly used parametric models are multivariate Gaussian distributions [159] for continuous features, binomial distributions for binary features, and multi-normal distributions for integer-valued (and categorical) features. A critical issue for Gaussian distributions is the assumption made about the covariance matrices. If the covariance matrices for different classes are assumed to be identical, then the Bayes plug-in rule, called Bayes normal-linear, provides a linear decision boundary. On the other hand, if the covariance matrices are assumed to be different, the resulting Bayes plug-in rule, which we call Bayes-normal-quadratic, provides a quadratic decision boundary. In addition to the commonly used maximum likelihood estimator of the covariance matrix, various regularization techniques [160] are available to obtain a robust estimate in small sample size situations and the leave-one-out estimator is available for minimizing the bias [161].

## 5. Performance of speech recognition systems:

The performance of speech recognition systems is usually specified in terms of accuracy and speed. Accuracy may be measured in terms of performance accuracy which is usually rated with word error rate (WER), whereas speed is measured with the real time factor. Other measures of accuracy include Single Word Error Rate (SWER) and Command Success Rate (CSR).

**Word Error Rate(WER):Word error rate** is a common metric of the performance of a speech recognition or machine translation system. The general difficulty of measuring performance lies in the fact that the recognized word sequence can have a different length from the reference word sequence (supposedly the correct one). The WER is derived from the Levenshtein distance, working at the word level instead of the phoneme level. This problem is solved by first aligning the recognized word sequence with the reference (spoken) word sequence using dynamic string alignment.

Word error rate can then be computed as:

$$WER = \frac{S + D + I}{N} \qquad \text{.......(9)}$$

where

- *S* is the number of substitutions,

- *D* is the number of the deletions,

- *I* is the number of the insertions,

- *N* is the number of words in the reference.

When reporting the performance of a speech recognition system, sometimes **word recognition rate (WRR)** is used instead:

$$WRR = 1 - WER = \frac{N - S - D - I}{N} = \frac{H - I}{N} \qquad \text{........(10)}$$

where

- *H* is N-(S+D), the number of correctly recognized words.

## 6. Literature Survey of speech recognition: ( year vise):

### 6.1 1920-1960s:

In the early 1920s machine recognition came into existence. The first machine to recognize speech to any significant degree commercially named, Radio Rex (toy) was manufactured in 1920[165].Research into the concepts of speech technology began as early as 1936 at Bell Labs. In 1939, Bell Labs demonstrated a speech synthesis machine (which simulates talking) at the World Fair in New York. Bell Labs later abandoned efforts to develop speech-simulated listening and recognition; based on an incorrect conclusion that artificial intelligence would ultimately be necessary for success.

The earliest attempts to devise systems for automatic speech recognition by machine were made in 1950s, when various researchers tried to exploit the fundamental ideas of acoustic phonetics. During 1950s[1], most of the speech recognition systems investigated spectral resonances during the vowel region of each utterance which were extracted from output signals of an analogue filter bank and logic circuits. In 1952, at Bell laboratories, Davis, Biddulph, and Balashek built a system for isolated digit recognition for a single speaker [2]. The system relied heavily on measuring spectral resonances during the vowel region of each digit. In an independent effort at RCA Laboratories in 1956, Olson and Belar tried to recognize 10 distinct syllables of a single talker, as embodied in 10 monosyllabic words [3]. The system again relied on spectral measurements (as provided by an analog filter bank) primarily during vowel regions. In 1959, at University College in England, Fry and Denes tried to build a phoneme recognizer to recognize four vowels and nine consonants [4]. They used a spectrum analyzer and a pattern matcher to make the recognition decision. A novel aspect of this research was the use of statistical information about allowable sequences of phonemes in English ( a rudimentary form of language syntax) to improve overall phoneme accuracy for words consisting of two or more phonemes. Another effort of note in this period was the vowel recognizer of Forgie and Forgie constructed at MIT Licoln laboratories in 1959 in which 10 vowels embedded in a /b/-vowel/t/ format were recognized in a speaker independent manner [5]. Again a Filter bank analyzer was used to provide spectral information and a time varying estimate of the vocal tract resonances was made to deicide which vowel was spoken.

### 6.2 1960-1970:

In the 1960s several fundamental ideas in speech recognition surfaced and were published. In the 1960s since computers were still not fast enough, several special purpose hardware were built [6]. However, the decade started with several Japanese laboratories entering the recognition arena and building special purpose hardware as part of their systems. On early Japanese system, described by Suzuki and Nakata of the Radio Research Lab in Tokyo, was a hardware vowel recognizer [7]. An elaborate filter bank spectrum analyzer was used along with logic that connected the outputs of each channel of the spectrum analyzer (in a weighted manner) to a vowel decision circuit, and majority decisions logic scheme was used to choose the spoken vowel. Another hardware effort in Japan was the work of Sakai and Doshita of kyoto University in 1962, who built a hardware phoneme recognizer [7]. A hardware speech segmented was used along with a zero crossing analysis of different regions of the spoken input to provide the recognition output. A third Japanese effort was the digit recognizer hardware of Nagata and coworkers at NEC Laboratories in 1963[8]. This effort was perhaps most notable as the initial attempt at speech recognition at NEC and led to a long and highly productive research program. One of the difficult problems of speech recognition exists in the non uniformity of time scales in speech events. In the 1960s three key research projects were initiated that have had major implications on the research and development of speech recognition for the past 20 years. The first of these projects was the efforts of Martin and his colleagues at RCA Laboratories, beginning in the late 1960s, to develop realistic solutions to the problems associated with non-uniformity of time scales in speech events. Martin developed a set of elementary time normalization methods, based on the ability to reliably detect speech starts and ends, that significantly reduce the variability of the recognition scores[9]. Martin ultimately developed the method and founded one of the first speech recognition companies, Threshold Technology, which was built, marketed and was sold speech recognition products. At about the same time, in the Soviet Union, Vintsyuk proposed the use of dynamic programming methods for time aligning a pair of speech utterances(generally known as Dynamic Time Warping(DTW) [10]),including algorithms for connected word recognition.. Although the essence of the concepts of dynamic time warping, as well as rudimentary versions of the algorithms for connected word recognition, were embodied in Vintsyuk's work, it was largely unknown in the West and did not come to light until the early 1980's; this was long after the more formal methods were proposed and implemented by others. At the same time in an independent effort in Japan Sakoe and Chiba at NEC Laboratories also started to use a dynamic Programming technique to solve the non uniformity problems[11].A final achievement of note in the 1960s was the pioneering research of Reddy in the field of continuous speech recognition by dynamic tracking of phonemes [12]. Reddy's research eventually spawned a long and highly successful speech recognition research program at Carnegie Mellon University (to which Reddy moved in the late 1960s) which, to this today, remains a world leader in continuous speech recognition systems.

### 6.3 1970-1980:

In the 1970s speech recognition research achieved a number of significant milestones. First the area of isolated word or discrete utterance recognition became a viable and usable technology based on fundamental studies by Velichko and Zagoruyko in Russia[13], Cakoe and Chiba in Japan[14], and Itakura in the united States. The Russian studies helped the advance use of pattern recognition ideas in speech recognition; the Japanese research showed how dynamic programming methods could be successfully applied; and Itakura's research showed how the ideas of linear predictive coding (LPC), which had already been successfully used in low bit rate speech coding, could be extended to speech recognition systems through the use of an appropriate distance measure based on LPC spectral parameters[15].Another milestone of the 1970s was the beginning of a longstanding, highly successful group effort in large vocabulary speech recognition at IBM in which researchers studied three distinct tasks over a period of almost two decades, namely the New Raleigh language [16] for simple database queries, the laser patent text language [17] for transcribing laser patents, and the office correspondent tasks called Tangora[18], for dictation of simple memos. Finally, at AT&T Bell Labs, researchers began a series of experiments aimed at making speech recognition systems that were truly speaker independent [19]. To achieve this goal a wide range of sophisticated clustering algorithms were used to determine the number of distinct patterns required to represent all variations of different words across a wide user population. This research has been refined over a decade so that the techniques for creating speaker independent patterns are now well understood and widely used. An ambitious speech understanding project was funded by the defence Advanced Research Projects Agencies(DARPA), which led to many seminal systems and technology[20]. One of the demonstrations of speech understanding was achieved by CMU in 1973 there Heresay I system was able to use semantic information to significantly reduce the number of alternatives considered by the recognizer.CMU's Harphy system[21] was shown to be able to recognize speech using a vocabulary of 1,011 words with reasonable accuracy. One of the particular contributions from the Harpy system was the concept of graph search, where the speech recognition language is represented as a connected network derived from lexical representations of words, with syntactical production rules and word boundary rules. The Harpy system was the first to take advantage of a finite state network (FSN) to reduce computation and efficiently determine the closest matching string. Other systems developed under the DARPA's speech understanding program included CMU's Hearsay II and BBN's HWIM (Hear what I Mean) systems[20]. The approach proposed by Hearsay II of using parallel asynchronous processes that simulate the component knowledge sources in a speech system was a pioneering concept. A global "blackboard" was used to integrate knowledge from parallel sources to produce the next level of hypothesis.

### 6.4 1980-1990:

Just as isolated word recognition was a key focus of research in the 1970s, the problems of connected word recognition was a focus of research in the 1980s. Here the goal was to create a robust system capable of recognizing a fluently spoken string of words(eg., digits) based on matching a concatenated pattern of individual words. Moshey J. Lasry has developed a feature-based speech recognition system in the beginning of 1980. Wherein his studies speech spectrograms of letters and digits[97].A wide variety of the algorithm based on matching a concatenated pattern of individual words were formulated and implemented, including the two level dynamic programming approach of Sakoe at Nippon Electric Corporation (NEC)[22],the one pass method of Bridle and Brown at Joint Speech Research Unit(JSRU) in UK[23], the level building approach of Myers and Rabiner at Bell Labs [24], and the frame synchronous level building approach of Lee and Rabiner at Bell Labs[25]. Each of these "optimal" matching procedures had its own implementation advantages, which were exploited for a wide range of tasks. Speech research in the 1980s was characterized by a shift in technology from template based approaches to statistical modeling methods especially the hidden Markov model approach [26,27]. Although the methodology of hidden Markov modeling (HMM) was well known and understood in a few laboratories(Primarily IBM, Institute for Defense Analyses (IDA), and Dargon systems), it was not until widespread publication of the methods and theory of HMMs, in the mid-1980, that the technique became widely applied in virtually, every speech recognition research laboratory in the world. Today, most practical speech recognition systems are based on the statistical frame work developed in the 1980s and their results, with significant additional improvements have been made in the 1990s.

### a) Hidden Markov Model(HMM):

HMM is one of the key technologies developed in the 1980s, is the hidden Markov model(HMM) approach [28,29,30]. It is a doubly stochastic process which as an underlying stochastic process that is not observable (hence the term hidden), but can be observed through another stochastic process that produces a sequence of observations. Although the HMM was well known and understood in a few laboratories (primarily IBM, Institute for Defense Analysis (IDA) and Dragon Systems), it was not until widespread publication of the methods and theory of HMMs in the mid-1980s that the technique became widely applied in virtually every speech recognition research laboratory in the world. In the early 1970s, Lenny Baum of

Princeton University invented a mathematical approach to recognize speech called Hidden Markov Modeling (HMM). The HMM pattern-matching strategy was eventually adopted by each of the major companies pursuing the commercialization of speech recognition technology (SRT).The U.S. Department of Defense sponsored many practical research projects during the '70s that involved several contractors, including IBM, Dragon, AT&T, Philips and others. Progress was slow in those early years.

### b) Neural Net:

Another "new" technology that was reintroduced in the late 1980s was the idea of applying neural networks to problems in speech recognition. Neural networks were first introduced in the 1950s, but they did not prove useful initially because they had many practical problems. In the 1980s however, a deeper understanding of the strengths and limitations of the technology was achieved, as well as, understanding of the technology to classical signal classification methods. Several new ways of implementing systems were also proposed [33,34,35].

### c) DARPA Program:

Finally, the 1980s was a decade in which a major impetus was given to large vocabulary, continuous speech recognition systems by the Defense Advanced Research Projects Agency (DARPA) community, which sponsored a large research program aimed at achieving high word accuracy for a 1000 word continuous speech recognition, database management task. Major research contributions resulted from efforts at CMU(notably the well known SPHINX system)[36], BBN with the BYBLOS system[37], Lincoln Labs[38], SRI[39], MIT[40], and AT&T Bell Labs[41]. The SPHINX system successfully integrated the statistical method of HMM with the network search strength of the earlier Harpy system. Hence, it was able to train and embed context dependent phone models in a sophisticated lexical decoding network. The DARPA program has continued into the 1990s, with emphasis shifting to natural language front ends to the recognizer and the task shifting to retrieval of air travel information. At the same time, speech recognition technology has been increasingly used within telephone networks to automate as well as enhance operator services.

### 6.5 1990-2000s:

In the 1990s a number of innovations took place in the field of pattern recognition. The problem of pattern recognition, which traditionally followed the framework of Bayes and required estimation of distributions for the data, was transformed into an optimization problem involving minimization of the empirical recognition error [42]. This fundamental paradigmatic change was caused by the recognition of the fact that the distribution functions for the speech signal could not be accurately chosen or defined and the Bayes decision theory becomes inapplicable under these circumstances. Fundamentally, the objective of a recognizer design should be to achieve the least recognition error rather than provide the best fitting of a distribution function to the given (known)data set as advocated by the Bayes criterion. This error minimization concept produced a number of techniques such as discriminative training and kernel based methods. As an example of discriminative training, the Minimum Classification Error(MCE) criterion was proposed along with a corresponding Generalized Probabilistic Descent(GPD) training algorithm to minimize an objective function which acts to approximate the error rate closely[43]. Another example was the Maximum Mutual Information (MMI) criterion. In MMI training, the mutual information between the acoustic observation and its correct lexical symbol averaged over a training set is maximized. Although this criterion is not based on a direct minimization of the classification error rate and is quite different from the MCE based approach, it is well founded in information theory and possesses good theoretical properties. Both the MMI and MCE can lead to speech recognition performance superior to the maximum likelihood based approach [43]. A key issue[82] in the design and implementation of speech recognition system is how to properly choose the speech material used to train the recognition algorithm. Training may be more formally defined as supervised learning of parameters of primitive speech patterns ( templates, statistical models, etc.,) used to characterize basic speech units ( e.g. word or subword units), using labeled speech samples in the form of words and sentences. It also discusses two methods for generating training sets. The first, uses a nondeterministic statistical method to generate a uniform distribution of sentences from a finite state machine represented in digraph form. The second method, a deterministic heuristic approach, takes into consideration the importance of word ordering to address the problem of co articulation effects that are necessary for good training. The two methods are critically compared.

### a) DARPA program:

The DARPA program continued into the 1990s, with emphasis shifting to natural language front ends to the recognizer. The central focus also shifted to the task of retrieving air travel information, the Air Travel Information Service (ATIS) task. Later the emphasis was expanded to a range of different speech-understanding application areas, in conjunction with a new focus on transcription of broadcast news (BN) and conversational speech. The Switchboard task is among the most challenging ones proposed by DARPA; in this task speech is conversational and spontaneous, with many instances of so-called disfluencies such as partial words, hesitation and repairs. The BN transcription technology was integrated with information extraction and retrieval technology, and many application systems, such as automatic voice document indexing and retrieval systems, were developed. A number of human language technology projects funded by DARPA in the 1980s and 1990s further accelerated the progress, as evidenced by many papers published in the proceedings of the DARPA Speech and Natural Language/Human Language Workshop. The research describes the development of activities for speech recognition that were conducted in the 1990s[83], at Fujitsu Laboratories Limited. Also, it is focused on extending the functions and performance of speech recognition technologies developed in the 1980s. Advnaces in small implementations of speech recognition, recognition of continuous speech, and recognition of speech in noisy environments, have been described.

### b) HMM :

A weighted hidden markov model HMM algorithm and a subspace projection algorithm are proposed in[109], to address the discrimination and robustness issues for HMM based speech recognition. Word models were constructed for combining phonetic and fenonic models[110] A new hybrid algorithm based on combination of HMM and learning vector were proposed in [111]. Learning Vector Quantisation[112] (LVQ) method showed an important contribution in producing highly discriminative reference vectors for classifying static patterns. The ML estimation of the parameters via FB algorithm was an inefficient method for estimating the parameters values of HMM. To over come this problem paper[113] proposed a corrective training method that minimized the number of errors of parameter estimation. A novel approach [114] for a hybrid connectionist HMM speech recognition system based on the use of a Neural Network as a vector qantiser. showed the important innovations in training the Neural Network. Next the Vector Quantization approach showed much of its significance in the reduction of Word error rate. MVA[115] method obtained from modified Maximum Mutual Information(MMI) is shown in this paper. Nam Soo Kim et.al., have presented various methods for estimating a robust output probability distribution(PD) in speech recognition based on the discrete Hidden Markov Model(HMM) in their paper[118].An extension of the viterbi algorithm[120] made the second order HMM computationally efficient when compared with the existing viterbi algorithm. In paper[123] a general stochastic model that encompasses most of the models proposed in the literature, pointing out similarities of the models in terms of correlation and parameter time assumptions, and drawing analogies between segment models and HMMs have been described. An alternative VQ[124] method in which the phoneme is treated as a cluster in the speech space and a Gaussian model was estimated for each phoneme. The results showed that the phoneme-based Gaussian modeling vector quantization classifies the speech space more effectively and significant improvements in the performance of the DHMM system have been achieved. The trajectory folding phenomenon in HMM model is overcome by using Continuous Density HMM which significantly reduced the Word Error Rate over continuous speech signal as been demonstrated by[125]. A new hidden Markov model[127] showed the integration of the generalized dynamic feature parameters into the model structure was developed and evaluated using maximum-likelihood (ML) and minimum-classification-error (MCE) pattern recognition approaches. The authors have designed the loss function for minimizing error rate specifically for the new model, and derived an analytical form of the gradient of the loss function that enables the implementation of the MCE approach. Authors extend[128] previously proposed quasi-Bayes adaptive learning framework to cope with the correlated continuous density hidden Markov models (HMM's) with Gaussian mixture state observation densities to implement the correlated mean vectors to be updated using successive approximation algorithm. Paper [130] investigates the use of Gaussian selection (GS) to increase the speed of a large vocabulary speech recognition system. The aim of GS is to reduce the load by selecting the subset of Gaussian component likelihoods for a given input vector, which also proposes new techniques for obtaining "good" Gaussian subsets or "shortlists" a novel framework of online hierarchical transformation of hidden Markov model (HMM) parameters[133], for adaptive speech recognition. Its goal is to incrementally transform (or adapt) all the HMM parameters to a new acoustical environment even though most of HMM units are unseen in observed adaptation data. The theoretical frame work[117] for Bayesian adaptive training of the parameters of discrete hidden markov model(DHMM) and semi continuous HMM(SCHMM) with Gaussian mixture state observation densities were proposed. The proposed MAP algorithms discussed in [117] are shown to be effective especially in the cases in which the training or adaptation data are limited.

### c) Robust speech recognition:
Various techniques were investigated to increase the robustness of speech recognition systems against the mismatch between training and testing conditions, caused by background noises, voice individuality, microphones, transmission channels, room reverberation, etc. Major techniques include, the maximum likelihood linear regression (MLLR) [44], the model decomposition [45], parallel model composition (PMC) [46], and the structural maximum a posteriori (SMAP) method [47] for robust speech recognition. The paper by Mazin G.Rahim et.al[116] presents a signal bias removal (SBR) method based on maximum likelihood estimation for the minimization of the undesirable effects which occur in telephone speech recognition system such as ambient noise, channel distortions etc.,. A maximum likelihood (ML) stochastic matching approach to decrease the acoustic mismatch between a test utterances, and a given set of speech models was proposed in [121] to reduce the recognition performance degradation caused by distortions in the test utterances and/or the model set. A new approach to an auditory model for robust speech recognition for noisy environments was proposed in [129] . The proposed model consists of cochlear bandpass filters and nonlinear operations in which frequency information of the signal is obtained by zero-crossing intervals. Compared with other auditory models, the proposed auditory model is computationally efficient, free from many unknown parameters, and able to serve as a robust front-end for speech recognition in noisy environments. Uniform distribution, is adopted to characterize the uncertainty of the mean vectors of the CDHMM's in [131].The paper proposed two methods, namely, a model compensation technique based on Bayesian predictive density and a robust decision strategy called Viterbi *Bayesian predictive classi fication* are studied. The proposed methods are compared with the conventional Viterbi decoding algorithm in speaker-independent recognition experiments on isolated digits and TI connected digit strings (TIDIGITS), where the mismatches between training and testing conditions are caused by: 1) additive Gaussian white noise, 2) each of 25 types of actual additive ambient noises, and 3) gender difference. A novel implementation of a mini-max decision rule for continuous density hidden Markov model-based robust speech recognition was proposed in [133]. By combining the idea of the mini-max decision rule with a normal Viterbi search, authors derive a recursive mini-max search algorithm, where the mini-max decision rule is repetitively applied to determine the partial paths during the search procedure.

### d) Noisy speech recognition:
Not much work has been done on noisy speech recognition in this decade. One of the important methods called minimum mean square error (MMSE) estimate of the filter log energies, introducing a significant improvement over existing algorithms were proposed by Adoram Erell and et.al. [98].A model based spectral estimation algorithm is derived that improves the robustness of SR system to additive noise. The algorithm is tailored for filter bank based system, where the estimation should seek to minimize the distortion as measured by the recognizer's distance [99]. Minor work has been done in the area of noisy robust speech recognition. A model based spectral estimation algorithm has been derived in [112] which improves the robustness' of the speech recognition system to additive noise. This algorithm is tailored for filter bank based systems where the estimation should seek to minimize the distortions as measured by the recognizers distance metric. The aim of this correspondence [126] is to present a robust representation of speech based on AR modeling of the causal part of the autocorrelation sequence. In noisy speech recognition, this new representation achieves better results than several other related techniques.

### 6.6. 2000-2009:
#### a) General:
Around 2000, a variational Bayesian (VB) estimation and clustering techniques were developed[71]. Unlike Maximum Likelihood, this VB approach is based on a posterior distribution of parameters. Giuseppe Richardi[73] have developed the technique to solve the problem of adaptive learning, in automatic speech recognition and also proposed active learning algorithm for ASR. In 2005, some improvements have been worked out on Large Vocabulary Continuous Speech Recognition[74] system on performance improvement. In 2007, the difference in acoustic features between spontaneous and read speech using a large scale speech data base i.e, CSJ have been analyzed[79]. Sadaoki Furui [81] investigated SR methods that can adapt to speech variation using a large number of models trained based on clustering techniques. In 2008, the authors[87] have explored the application of Conditional Random Field(CRF) to combine local posterior estimates provided by multilayer perceptions corresponding to the frame level prediction of phone and phonological attributed classes. De-wachter et.al.[100], attempted to over-come the time dependencies, problems in speech recognition by using straight forward template matching method. Xinwei Li et.al.[105], proposed a new optimization method i.e., semi definite programming(SDP) to solve the large margin estimation(LME) problem of continuous density HMM(CDHMM) in speech recognition. Discriminate training of acoustic models for speech recognition was proposed under Maximum mutual information(MMI)[107]. Around 2007 Rajesh M.Hegde et.al, [106], proposed an alternative method for processing the Fourier transform phase for extraction speech features, which process the group delay feature(GDF) that can be directly computed for the speech signal.

### b) DARPA program:

The Effective Affordable Reusable Speech-to-Text (EARS) program was conducted to develop speech-to-text (automatic transcription) technology with the aim of achieving substantially richer and much more accurate output than before. The tasks include detection of sentence boundaries, fillers and disfluencies. The program was focusing on natural, unconstrained human speech from broadcasts and foreign conversational speech in multiple languages. The goal was to make it possible for machines to do a much better job of detecting, extracting, summarizing and translating important information, thus enabling humans to understand what was said by reading transcriptions instead of listening to audio signals [48, 49].

### c) Spontaneous speech recognition:

Although read speech and similar types of speech, e.g. news broadcasts reading a text, can be recognized with accuracy higher than 95% using state-of-the-art of speech recognition technology, and recognition accuracy drastically decreases for spontaneous speech. Broadening the application of speech recognition depends crucially on raising recognition performance for spontaneous speech. In order to increase recognition performance for spontaneous speech, several projects have been conducted. In Japan, a 5-year national project "Spontaneous Speech: Corpus and Processing Technology" was conducted [50]. A world-largest spontaneous speech corpus, "Corpus of Spontaneous Japanese (CSJ)" consisting of approximately 7 millions of words, corresponding to 700 hours of speech, was built, and various new techniques were investigated. These new techniques include flexible acoustic modeling, sentence boundary detection, pronunciation modeling, acoustic as well as language model adaptation, and automatic speech summarization [51]. The three analyses on the effects of spontaneous speech on continuous speech recognition performance are described in [93] viz., (1) spontaneous speech effects significantly degrade recognition performance, (2)

*fluent* spontaneous speech yields word accuracies equivalent to read speech, and (3) using spontaneous speech training data. These can significantly improve the performance for recognizing spontaneous speech. It is concluded that word accuracy can be improved by explicitly modeling spontaneous effects in the recognizer, and by using as much spontaneous speech training data as possible. Inclusion of read speech training data, even within the task domain, does not significantly improve performance.

### d) Robust speech recognition:

To further increase the robustness of speech recognition systems, especially for spontaneous speech, utterance verification and confidence measures, are being intensively investigated [52]. In order to have intelligent or human-like interactions in dialogue applications, it is important to attach to each recognized event a number that indicates how confidently the ASR system can accept the recognized events. The confidence measure serves as a reference guide for a dialogue system to provide an appropriate response to its users. To detect semantically, significant parts and reject irrelevant portions in spontaneous utterances, a detection based approach has recently been investigated [53]. The combined recognition and verification strategy work well especially for ill-formed utterances. In order to build acoustic models more sophisticated than conventional HMMs, the dynamic Bayesian network has recently been investigated [54]. Around 2000, a QBPC[56], systems were developed to find the unknown and mismatch between training and testing conditions. A DCT fast subspace techniques[60] has been proposed to approximate the KLT for autoregressive progress. A novel implementation of a mini-max decision rule for continuous density HMM-based Robust speech recognition is developed by combining the idea of mini-max decision rule with a normal viterbi search. Speech signal modeling techniques well suited to high performance and robust isolated word recognition have been contributed[61,63]. The first robust Large vocabulary continuous speech recognition that uses syllable-level acoustic unit of LVCSR on telephone bandwidth speech is described in [64]. In 2003, a novel regression based Bayesian predictive classification(LRBPC[69]) was developed for speech Hidden markov model. Walfgang Rchichal[62] has described the methods of improving the robustness and accuracy of the acoustic modeling using decision tree based state tying. Giuluva Garau et.al.[85], investigated on Large vocabulary continuous speech recognition. Xiong Xiao[92] have shown a novel technique that normalizes the modulation spectra of speech signal. Kernel based nonlinear predictive coding[101] procedure, that yields speech features which are robust to non-stationary noise contaminated speech signal. Features maximally in sensitive to additive noise are obtained by growth transformation of regression functions that span a reproducing a kernel Hilbert space (RKHS). Soundararajan [103] proposed a supervised approach using regression trees to learn non linear transformation of the uncertainty from the linear spectral domain to the cepstral domain. Experiments are conducted on Aurora-4 Database.

### e) Multimodal speech recognition:

Humans use multimodal communication when they speak to each other. Studies in speech intelligibility have shown that having both visual and audio information increases the rate of successful transfer of information, especially when the message is complex or when communication takes place in a noisy environment. The use of the visual face information, particularly lip information, in speech recognition has been investigated, and results show that using both types of information gives better recognition performances than using only the audio or only the visual information, particularly, in noisy environment. Jerome R., have developed Large Vocabulary Speech Recognition with Multi-span Statistical Language Models [55] and the work done in this paper characterizes the behavior of such multi span modeling in actual recognition. A novel subspace modeling is presented in [84], including selection approach for noisy speech recognition. In subspace modeling, authors have developed a factor analysis representation of noisy speech i.e., a generalization of a signal subspace representation. They also explored the optimal subspace selection via solving the hypothesis test problems. Subspace selection via testing the correlation of residual speech, provides high recognition accuracies than that of testing the equivalent eigen-values in the minor subspace. Because of the environmental mismatch between training and test data severely deteriorates recognition performance. Jerome R. et.al.[55], have contributed large vocabulary speech recognition with multi-span statistical language model.

### f) Modeling Techniques:

Eduardo et.al.[56], introduced a set of acoustic modeling and decoding techniques for utterance verification(UV) in HMM based Continuous Speech Recognition .Lawerence K et.al.[58], discuss regarding HMM models for Automatic speech recognition which rely on high dimension feature vectors for summarizing the short time properties of speech. These have been achieved using some parameters choosen in two ways, namely i) to maximize the likelihood of observed speech signals, or ii) to minimize the number of classification errors. Dat Tat Tran[75] have proposed various models namely, i) the FE-HMM,NC-FE-HMM,FE-GMM,NC-FE-GMM,FE-VQ and NC-FE-VQ in the FE approach, ii) the FCM-HMM, NC-FCM-HMM,FCM-GMM and NC-FCM-GMM in the FCM approach and iii) the hard HMM and GMM as the special models of both FE and FCM approaches for speech recognition. A new statistical approach namely the probabilistic union model for Robust speech recognition involving partial, unknown frequency[67] band corruption are introduced by Ji Ming et.al. Jen Tzung et.al.[69], have surveyed a series of model selection approaches with a presentation of a novel predictive information criterion for HMM selection. Yang Liu et.al.[95], have shown that in a metadata detection scheme in speech recognition discriminative models outperform generative than predominant HMM approaches. Alba Sloin et.al. have presented a discriminative training algorithm, that uses support vector machines(SVM) to improve the classification of discrete and continuous output probability hidden markov models. The algorithm presented in the paper[119] uses a set of maximum likelihood (ML) trained HMM models as a baseline system, and an SVM training scheme to rescore the results of the baseline HMMs. The experimental results given in that paper reduces the error rate significantly compared to standard ML training. Paper[140] presents a discriminative training algorithm that uses support vector machines(SVMs) to improve the classification of discrete and continuous output probability hidden markov models(HMMs). The algorithm uses a set of maximum likelihood (ML) trained HMM models as a baseline system, and an SVM training scheme to rescore the results of the baseline HMMs. Paper[142], proposes a Fuzzy approach to the hidden Markov model (HMM) method called the fuzzy HMM for speech and speaker recognition as an application of fuzzy expectation maximizing algorithm in HMM. This fuzzy approach can be applied to EM-style algorithms such as the Baum- Welch algorithm for hidden Markov models, the EM algorithm for Gaussian mixture models in speech and speaker recognition. Equation and how estimation of discrete and continuous HMM parameters based on this two algorithm is explained and performance of two methods of speech recognition for one hundred words is surveyed . This paper showed better results for the fuzzy HMM, compared with the conventional HMM**.** A novel method to estimate continuous-density hidden Markov model (CDHMM) for speech recognition [143] is, according to the principle of maximizing the minimum multi-class separation margin. The approach is named large margin HMM. First, they showed that this type of large margin HMM estimation problem can be formulated as a constrained mini-max optimization problem. Second, they propose to solve this constrained mini-max optimization problem by using a penalized gradient descent algorithm, where the original objective function, i.e., minimum margin, is approximated by a differentiable function and the constraints are cast as penalty terms in the objective function. Ultimately paper showed that the large margin training method yields significant recognition error rate reduction even on top of some popular discriminative training methods.

In the work[145], techniques for recognizing phonemes automatically by using Hidden Markov Models (HMM) were proposed. The features input to the HMMs will be extracted from a single phoneme directly rather than from a string of phonemes forming a word. Also feature extraction techniques are compared to their performance in phoneme-based recognition systems. They also describe a pattern recognition approach developed for continuous speech recognition. Modeling dynamic structure of speech[146] is a novel paradigm in speech recognition research within the generative modeling framework, and it offers a potential to overcome limitations of the current hidden Markov modeling approach. Analogous to structured language models where syntactic structure is exploited to represent long-distance relationships among words [5], the structured speech model described in this paper make use of the dynamic structure in the hidden vocal tract resonance space to characterize long-span contextual influence among phonetic units. The paper[147], discusses two novel HMM based techniques that segregate a speech segment from its concurrent background. The first method can be reliably used in clean environments while the second method, which makes use of the wavelets denoising technique, is effective in noisy environments. These methods have been implemented and they showed the superiority over other popular techniques, thus, indicating that they have the potential to achieve greater levels of accuracy in speech recognition rates. Paper[162], is motivated by large margin classifiers in machine learning. It proposed a novel method to estimate continuous-density hidden Markov model (CDHMM) for speech recognition according to the principle of maximizing the minimum multi-class separation margin. The approach is named as large margin HMM. First, it shows this type of large margin HMM estimation problem can be formulated as a constrained mini-max optimization problem. Second, it proposes to solve this constrained mini-max optimization problem by using a penalized gradient descent algorithm, where the original objective function, i.e., minimum margin, is approximated by a differentiable function and the constraints are cast as penalty terms in the objective function. The new training method is evaluated in the speaker-independent isolated E-set recognition and the TIDIGITS connected digit string recognition tasks. Experimental results clearly show that the large margin HMMs consistently outperform the conventional HMM training methods. It has been consistently observed that the large margin training method yields significant recognition error rate reduction even on top of some popular discriminative training methods. Despite their known weaknesses, hidden Markov models (HMMs) have been the dominant technique for acoustic modeling in speech recognition for over two decades. Still, the advances in the HMM framework have not solved its key problems: it discards information about time dependencies and is prone to overgeneralization. Paper[163], has attempted to overcome the above problems by relying on straightforward template matching. It showed the decrease in word error rate with 17% compared to the HMM results. In automatic speech recognition, hidden Markov models (HMMs) are commonly used for speech decoding, while switching linear dynamic models (SLDMs) can be employed for a preceding model-based speech feature enhancement. These model types are combined[164] in order to obtain a novel iterative speech feature enhancement and recognition architecture. It is shown that speech feature enhancement with SLDMs can be improved by feeding back information from the HMM to the enhancement stage. Two different feedback structures are derived. In the first, the posteriors of the HMM states are used to control the model probabilities of the SLDMs, while in the second they are employed to directly influence the estimate of the speech feature distribution. Both approaches lead to improvements in recognition accuracy both on the AURORA-2 and AURORA-4 databases compared to non-iterative speech feature enhancement with SLDMs. It is also shown that a combination with uncertainty decoding further enhances performance.

## g) Noisy speech recognition:

In 2008[89], a new approach for speech feature enhancement in the log spectral domain for noisy speech recognition is presented. A switching linear dynamic model (SLDM) is explored as a parametric model for the clean speech distribution. The results showed that the new SLDM approach can further improve the speech feature enhancement performance in terms of noise robust recognition accuracy. Jen-Tzung et.al.[69], present a novel subspace modeling and selection approach for noisy speech recognition. Jianping Dingebal [89] have presented a new approach for speech feature enhancement in the large spectral domain for NSR. Xiaodong[108] propose a novel approach which extends the conventional GMHMM by modeling state emission(mean and variance) as a polynomial function of a continuous environment dependent variable. This is used to improve the recognition performance in noisy environments by using multi condition training. Switching Linear dynamical system(SLDC)[102], is a new model that combines both the raw speech signal and the noise was introduced in the year 2008. This was tested using isolated digit utterance corrupted by Gaussian noise. Contrary to Autoregressive HMMs,SLDC's outperforms a state of the art feature based HMM. Mark D.Skowronski[104], proposed echo state network classifier by combining ESN with state machine frame work for noisy speech recognition. In the paper[144], authors propose a novel approach which extends the conventional Gaussian mixture hidden Markov model (GMHMM) by modeling state emission parameters (mean and variance) as a polynomial function of a continuous environment-dependent variable. At the recognition time, a set of HMMs specific to the given value of the environment variable is instantiated and used for recognition. The maximum-likelihood (ML) estimation of the polynomial functions of the proposed variable-parameter GMHMM is given within the expectation-maximization (EM) framework.

## h) Data driven approach:

A new approach [134], for deriving compound words from a training corpus was proposed. The motivation for making compound words is because under some assumptions, speech recognition errors occur less frequently in longer words were discussed along with the accurate modeling . They have also introduced a measure based on the product between the direct and the reverse bi-gram probability of a pair of words for finding candidate pairs in order to create compound words. Paper[135] surveys a series of model selection approaches and presents a novel *predictive information criterion* (PIC) for hidden Markov model (HMM) selection. The approximate Bayesian using Viterbi approach is applied for PIC selection of the best HMMs providing the largest prediction information for generalization of future data. Authors have developed a top-down *prior/posterior propagation* algorithm for estimation of structural hyper-parameters and they have showed the evaluation of continuous speech recognition(data driven) using decision tree HMMs, the PIC criterion outperforms ML and MDL criteria in building a compact tree structure with moderate tree size and higher recognition rate. A method of compensating for nonlinear distortions in speech representation caused by noise was proposed which is based on the histogram equalization. Paper[138], introduces the data driven signal decomposition method based on the empirical mode decomposition(EMD) technique. The decomposition process uses the data themselves to derive the base function in order to decompose the one-dimensional signal into a finite set of intrinsic mode signals. The novelty of EMD is that the decomposition does not use any artificial data windowing which implies fewer artifacts in the decomposed signals. The results show that the method can be effectively used in analyzing non-stationary signals.

## 7. Speech Databases:

Speech databases have a wider use in Automatic Speech Recognition. They are also used in other important applications like, Automatic speech synthesis, coding and analysis including speaker and language identification and verification. All these applications require large amounts of recoded database. Different types of databases that are used for speech recognition applications are discussed along with its taxonomy.

**Taxonomy of Existing Speech Databases:**

The intra-speaker and inter-speaker variability are important parameters for a speech database. Intra-speaker variability is very important for speaker recognition performance. The intra-speaker variation can originate from a variable speaking rate, changing emotions or other mental variables, and in environment noise. The variance brought by different speakers is denoted inter-speaker variance and is caused by the individual variability in vocal systems involving source excitation, vocal tract articulation, lips and/or nostril radiation. If the inter-speaker variability dominates the intra-speaker variability, speaker recognition is feasible. Speech databases are most commonly classified into single-session and multi-session. Multi-session databases allow estimation of temporal intra-speaker variability. According to the acoustic environment, databases are recorded either in noise free environment, such as in the sound booth, or with office/home noise. Moreover, according to the purpose of the databases, some corpora are designed for developing and evaluating speech recognition, for instance TIMIT, and some are specially designed for speaker recognition, such as SIVA, Polycost and YOHO. Many databases were recorded in one native language of recording subjects; however there are also multi-language databases with non-native language of speakers, in which case, the language and speech recognition become the additional use of those databases.

**Main database characteristics:**

Table-6 represents the characteristics of main databases used in speech recognition.

**Abbreviations:**

QR: Quiet Room
Ofc: Office
RF:Radio Frequency

Table 6: Database Characteristics:

| Name | No. speakers | No. units | Speech style | Recording environment | SR kHz | Transcription based on |
|---|---|---|---|---|---|---|
| TI Digits | 326 | >2500 numbers | Reading | QR | 20 | Word |
| TIMIT | 630 | 6300 sentences | Reading | QR | 16 | Phones |
| NTIMIT | 630 | 6300 sentences | Reading | Tel | 8 | Phones |
| RM1 | 144 | 15024 sentences | Reading | QR | 20 | Sentence |
| RM2 | 4 | 10608 sentences | Reading | QR | 20 | Sentence |
| ATIS0 | 36 | 10722 utterances | Reading Spont | Ofc. | 16 | Sentence |
| Switch Board(Credit card) | 69 | 35 dialogues | Conv Spont | Tel | 8 | Word |
| TI-46 | 16 | 19,136 isol.words | Reading | QR | 16 | Sentence |
| Switch Board(complete) | 550 | 2500 dialogues | Conv Spont | Tel | 8 | Word |
| ATC | 100 | 30000 utterances | Spont | RF | 8 | Sentence |
| ATIS2 | 351 | 12000 utterances | Spont | Ofc. | 16 | Sentence |

### 7.1. Resource Management Complete Set 2.0:

The DARPA Resource Management Continuous Speech Corpora (RM) consists of digitized and transcribed speech for use in designing and evaluating continuous speech recognition systems. There are two main sections, often referred to as RM1 and RM2. RM1 contains three sections, Speaker-Dependent (SD) training data, Speaker-Independent (SI) training data and test and evaluation data. RM2 has an additional and larger SD data set, including test material. All RM material consists of read sentences modeled after a naval resource management task. The complete corpus contains over 25,000 utterances from more than 160 speakers representing a variety of American dialects. The material was recorded at 16KHz, with 16-bit resolution, using a Sennheiser HMD-414 headset microphone. All discs conform to the ISO-9660 data format.

### 7.1.1. Resource Management SD and SI Training and Test Data (RM1):

The Speaker-Dependent (SD) Training Data contains 12 subjects, each reading a set of 600 "training sentences," two "dialect" sentences and ten "rapid adaptation" sentences, for a total of 7,344 recorded sentence utterances. The 600 sentences designated as training cover 97 of the lexical items in the corpus. The Speaker-Independent (SI) Training Data contains 80 speakers, each reading two "dialect" sentences plus 40 sentences from the Resource Management text corpus, for a total of 3,360 recorded sentence utterances. Any given sentence from a set of 1,600 Resource Management sentence texts were recorded by two subjects, while no sentence was read twice by the same subject.

RM1 contains all SD and SI system test material used in 5 DARPA benchmark tests conducted in March and October of 1987, June 1988 and February and October 1989, along with scoring and diagnostic software and documentation for those tests. Documentation is also provided outlining the use of the Resource Management training and test material at CMU in development of the SPHINX system. Example output and scored results for state-of-the-art speaker-dependent and speaker-independent systems (i.e. the BBN BYBLOS and CMU SPHINX systems) for the October 1989 benchmark tests are included.

### 7.1.2. Extended Resource Management Speaker-Dependent Corpus (RM2):

This set forms a speaker-dependent extension to the Resource Management (RM1) corpus. The corpus consists of a total of 10,508 sentence utterances (two male and two female speakers each speaking 2,652 sentence texts). These include the 600 "standard" Resource Management speaker-dependent training sentences, two dialect calibration sentences, ten rapid adaptation sentences, 1,800 newly-generated extended training sentences, 120 newly-generated development-test sentences and 120 newly-generated evaluation-test sentences. The evaluation-test material on this disc was used as the test set for the June 1990 DARPA SLS Resource Management Benchmark Tests (see the Proceedings). The RM2 corpus was recorded at Texas Instruments. The NIST speech recognition scoring software originally distributed on the RM1 "Test" Disc was adapted for RM2 sentences.

### 7.2. TIMIT:

**TIMIT** is a corpus of phonemically and lexically transcribed speech of American English speakers of different sexes and dialects. Each transcribed element has been delineated in time.

TIMIT was designed to further acoustic-phonetic knowledge and automatic speech recognition systems. It was commissioned by DARPA and worked on by many sites, including Texa Instrument (TI) and Masachusetts Institute of Technology (MIT), hence the corpus' name. There is also a telephone bandwidth version called NTIMIT (Network TIMIT). The TIMIT corpus of read speech is designed to provide speech data for acoustic-phonetic studies and for the development and evaluation of automatic speech recognition systems. Although it was primarily designed for speech recognition, it is also widely used in speaker recognition studies, since it is one of the few databases with a relatively large number of speakers. It is a single-session database recorded in a sound booth with fixed wideband headset. TIMIT contains broadband recordings of 630 speakers of eight major dialects of American English, each reading ten phonetically rich sentences. The TIMIT corpus includes time-aligned orthographic, phonetic and word transcriptions as well as a 16-bit, 16kHz speech waveform file for each utterance.

Corpus design was a joint effort among the Massachusetts Institute of Technology (MIT), SRI International (SRI) and Texas Instruments, Inc. (TI). There are numerous corpora for speech recognition. The most popular bases are: TIMIT and its derivatives, Polycost, and YOHO.

### 7.2.1. TIMIT and Derivatives:

The derivatives of TIMIT are: CTIMIT, FFMTIMIT, HTIMIT, NTIMIT, VidTIMIT. They were recorded by playing different recording input devices, such as telephone handset lines and cellular telephone handset, etc. TIMIT and most of the derivatives are single-session, and are thus not optimal for evaluating speaker recognition systems because of lack of intra-speaker variability. VidTIMIT is an exception, being comprised of video and corresponding audio recordings of 43 subjects. It was recorded into three sessions with around one week delay between each session. It can be useful for research involving automatic visual or audio-visual speech recognition or speaker verification.

### 7.3 TI46 database:

The TI46 corpus was designed and collected at Texas Instruments(TI). The speech was produced by 16 speakers, 8 females and 8 males, labeled f1-f8 and m1-m8 respectively, consisting of two vocabularies TI-20 and TI-alphabet. The TI-20 vocabulary contains the ten digits from 0 to 9 and ten command words: enter, erase, go, help, no, robot, stop, start, and yes. The TI alphabet vocabulary contains the names of the 26 letters of the alphabet from a to z. For each vocabulary item each speaker produced 10 tokens in a single training session and another two tokens in each of 8 testing sessions.

### 7.4 SWITCHBOARD:

SWITCHBOARD is a large multi-speaker corpus of telephone conversations. Although designed to support several types of speech and language research, its variety of speakers, speech data, telephone handsets, and recording conditions make SWITCHBOARD a rich source for speaker verification experiments of several kinds. Collected at Texas Instruments with funding from ARPA, SWITCHBOARD includes about 2430 conversations averaging 6 minutes in length; in other terms, over 240 hours of recorded and transcribed speech, about 3 million words, spoken by over 500 speakers of both sexes from every major dialect of American English. The data is 8 kHz, 8-bit mu-law encoded, with the two channels inter-leaved in each audio. In addition to its volume, SWITCHBOARD has a number of unique features contributing to its value for telephone-based speaker identification technology development. SWITCHBOARD was collected without human intervention, under computer control. From human factors perspective, automation guards against the intrusion of experimenter bias, and guarantees a degree of uniformity throughout the long period of data collection. The protocols were further intended to elicit natural and spontaneous speech by the participants.

Each transcript is accompanied by a time alignment i.e, which estimates the beginning time and duration of each word in the transcript in centi-seconds. The time alignment was accomplished with supervised phone-based speech recognition, as described by Wheatley et al. The corpus is therefore capable of supporting not only purely text-independent approaches to speaker verification, but also those which make use of any degree of knowledge of the text, including phonetics. SWITCHBOARD has both depth and breadth of coverage for studying speaker characteristics. Forty eight people participated 20 times or more; this yields at least an hour of speech, enough for extensive training or modeling and for repeated testing with unseen material. Hundreds of others participated ten times or less, providing a pool large enough for many open-set experiments. The participants demographics, as well as the dates, times, and other pertinent information about each phone call, are recorded in relational database tables. Except for personal information about the callers, these tables are included with the corpus. The volunteers who participated provided information relevant to studies of voice, dialect, and other aspects of speech style, including age, sex, education, current residence, and places of residence during formative years. The exact time and the area code of origin of each call is provided, as well as a means of telling which calls by the same person came from different telephones.

### 7.5. Air Travel Information System(ATIS):

The ATIS database is commonly used for the evaluation of word error performances of the Automatic Speech Recognition. ATIS is based on a realistic application environment and is a good simulation of spontaneous conversation.

### 8. Summary of the technology progress:

In the last 60 years, especially in the last three decades, research in speech recognition has been intensively carried out world wide, spurred on by advances in signal processing algorithms, architectures and hardware. The technological progress in the 60 years can be summarized in the table 7[137].

Table 7: Summary of the technological progress in the last 60 years

| Sl.No. | Past | Present(new) |
|---|---|---|
| 1) | Template matching | Corpus-based statistical modeling, e.g. HMM and n grams |
| 2) | Filter bank/spectral resonance | Cepstral features, Kernel based function, group delay functions |
| 3) | Heuristic time normalization | DTW/DP matching |
| 4) | "Distance"-based methods | Likelihood based methods |
| 5) | Maximum likelihood approach | Discriminative approach e.g. MCE/GPD and MMI |
| 6) | Isolated word recognition | Continuous speech recognition, |
| 7) | Small vocabulary | Large vocabulary |
| 8) | Context Independent units | Context dependent units |
| 9) | Clean speech recognition | Noisy/telephone speech recognition |
| 10) | Single speaker recognition | Speaker-independent/adaptive recognition |
| 11) | Monologue recognition | Dialogue/Conversation recognition |
| 12) | Read speech recognition | Spontaneous speech recognition |
| 13) | Single modality(audio signal only) | Multimodal(audio/visual)speech recognition |
| 14) | Hardware recognizer | Software recognizer |
| 15) | Speech signal is assumed as quasi –stationary in the traditional approaches. The feature vectors are extracted using FFT and wavelet methods etc.,. | Data driven approach does not posses this assumption i.e. signal is treated as nonlinear and non-stationary. In this features are extracted using Hilbert Haung Transform using IMFs.[141] |

### 9. Gap between machine and human speech recognition:

What we know about human speech processing is still very limited, and we have yet to witness a complete and worthwhile unification of the science and technology of speech. In 1994,

Moore [96] presented the following 20 themes which is believed to be an important to the greater understanding of the nature of speech and mechanisms of speech pattern processing in general:

- How important is the communicative nature of speech?
- Is human-human speech communication relevant to human machine communication by speech?
- Speech technology or speech science?(How can we integrate speech science and technology).
- Whither a unified theory?
- Is speech special?
- Why is speech contrastive?
- Is there random variability in speech?
- How important is individuality?
- Is disfluency normal?
- How much effort does speech need?
- What is a good architecture (for speech processes)?
- What are suitable levels of representation?
- What are the units?
- What is the formalism?
- How important are the physiological mechanisms?
- Is time-frame based speech analysis sufficient?
- How important is adaptation?
- What are the mechanisms for learning?
- What is speech good for?
- How good is speech.

After more than 10 years, we still do not have clear answers to these 20 questions.

## 10. Discussions and Conclusions:

Speech is the primary, and the most convenient means of communication between people. Whether due to technological curiosity to build machines that mimic humans or desire to automate work with machines, research in speech and speaker recognition, as a first step toward natural human-machine communication, has attracted much enthusiasm over the past five decades. we have also encountered a number of practical limitations which hinder a widespread deployment of application and services. In most speech recognition tasks, human subjects produce one to two orders of magnitude less errors than machines. There is now increasing interest in finding ways to bridge such a performance gap. What we know about human speech processing is very limited. Although these areas of investigations are important the significant advances will come from studies in acoustic-phonetics, speech perception, linguistics, and psychoacoustics. Future systems need to have an efficient way of representing, storing, and retrieving "knowledge" required for natural conversation. This paper attempts to provide a comprehensive survey of research on speech recognition and to provide some year wise progress to this date. Although significant progress has been made in the last two decades, there is still work to be done, and we believe that a robust speech recognition system should be effective under full variation in: environmental conditions, speaker variability's etc. Speech Recognition is a challenging and interesting problem in and of itself. We have attempted in this paper to provide a comprehensive cursory, look and review of how much speech recognition technology progressed in the last 60 years. Speech recognition is one of the most integrating areas of machine intelligence, since, humans do a daily activity of speech recognition. Speech recognition has attracted scientists as an important discipline and has created a technological impact on society and is expected to flourish further in this area of human machine interaction. We hope this paper brings about understanding and inspiration amongst the research communities of ASR.

### ACKNOWLEDGMENTS

The authors remain thankful to Dr.G.Krishna (Rtd. Prof. IISc, Bangalore)and Dr.M.Narashima Murthuy, Prof. & Chairman, Dept. of CS & Automation, IISc., Bangalore, for their useful discussions and suggestions during the preparation of this technical paper.